\documentclass[letterpaper]{article} 
\usepackage{aaai2026}  
\usepackage{times}  
\usepackage{helvet}  
\usepackage{courier}  
\usepackage[hyphens]{url}  
\usepackage{graphicx} 
\usepackage{natbib}  
\usepackage{multicol}
\usepackage{multirow}
\usepackage{dblfloatfix}

\usepackage{tabularx}
\usepackage{booktabs}
\usepackage{algorithm2e}
\usepackage{subcaption}
\usepackage{caption} 
\usepackage{tikz}
\usepackage{tikz-network}
\usetikzlibrary{arrows.meta}
\usetikzlibrary{positioning,fit,backgrounds}
\usetikzlibrary{decorations.markings}

\usepackage{algorithm}
\usepackage{algorithmic}

\title{\vspace{-1em} Small LLMs with Expert Blocks Are Good Enough for Hyperparamter Tuning}
\author{
    Om Naphade\equalcontrib\textsuperscript{\rm 1}, Saksham Bansal\equalcontrib\textsuperscript{\rm 2}, Parikshit Pareek\textsuperscript{\rm 3}$^\ddagger$
}
\affiliations{
 \textsuperscript{\rm 1} Department of Physics,
  \textsuperscript{\rm 2} Department of Mechanical Engineering,
    \textsuperscript{\rm 3}Department of Electrical Engineering\\
    Indian Institute of Technology Roorkee (IIT Roorkee),
    Uttarakhand, India.\\
    \rm{om\_nn@ph.iitr.ac.in, saksham\_b@me.iitr.ac.in, pareek@ee.iitr.ac.in}
}

\begin{document}

\maketitle

\begin{abstract}
Hyper-parameter Tuning (HPT) is a necessary step in machine learning (ML) pipelines but becomes computationally expensive and opaque with larger models. Recently, Large Language Models (LLMs) have been explored for HPT, yet most rely on models exceeding 100 billion parameters. We propose an Expert Block Framework for HPT using Small LLMs. At its core is the Trajectory Context Summarizer (TCS), a deterministic block that transforms raw training trajectories into structured context, enabling small LLMs to analyze optimization progress with reliability comparable to larger models. Using two locally-run LLMs (phi4:reasoning14B and qwen2.5-coder:32B) and a 10-trial budget, our TCS-enabled HPT pipeline achieves average performance within \textbf{~0.9 percentage points} of GPT-4 across six diverse tasks.
\end{abstract}
\begin{links}
    \link{Code}{https://github.com/PSquare-Lab/LLM-TCS-HPT}
\end{links}

\section{Introduction}
Hyperparameter tuning (HPT) is a critical step in achieving optimal model performance, but conventional approaches such as Bayesian Optimization and AutoML suffer from high computational cost, complex setup, and limited interpretability \cite{optuna2019,automl2019}. Recent advances in Large Language Models (LLMs) have opened new possibilities for addressing these limitations by \emph{asking LLMs to run HPT trials, and find best Hyperparamters within given time/computation budget}. Recent work has shown that LLMs can outperform traditional search methods by leveraging chain-of-thought reasoning and extended optimization horizons \cite{chainofthought}, while frameworks such as AgentHPO \cite{AgentHPO} and OptiMindTune \cite{madiraju2025optimindtunemultiagentframeworkintelligent} demonstrate the potential of multi-agent and structured reasoning approaches for HPT. However, these methods rely on large, resource-intensive models such as GPT-4, Gemini 1.5, raising questions of cost, accessibility, and scalability. 

In this work we attempt to answer the question—\emph{Can an HPT-specific deterministic expert block help small LLMs perform at par with large LLMs for the HPT problem?} To answer this question, we introduce a Trajectory Context Summarizer (TCS) which deterministically ingests trial artifacts (trial hyperparameters, per-epoch results for the current run, and aggregated history of prior runs) and emits a compact, machine- and human-readable state report (current status, latest experiment summary, per-parameter history, and comparative effect of recent changes). This structured report reduces token use and noisy signals, so small LLMs can reason about optimization progress and propose reliable hyperparameter updates.

\newcommand{\fix}{\marginpar{FIX}}
\newcommand{\new}{\marginpar{NEW}}

\section{Related Works}
Hyperparameters are configuration settings defined before training a machine learning model that control its learning process and performance (e.g., learning rate, number of layers, batch size). As the performance of machine learning models is sensitive to these parameters, hyperparameter tuning (HPT) is performed to obtain suitable hyperparameters. Standard HPT methods includes grid search, random search, Bayesian optimization (BO), and end-to-end AutoML pipelines, but these approaches become computationally heavy when model evaluations are expensive and the pipeline is opaque or rigid \cite{shahriari2015taking,automl2019}. Below we briefly review these methods to contextualize proposed approach for HPT.

\subsection{Bayesian Optimization} Bayesian optimization (BO) is a black-box optimization method which does not require gradients of the model with respect to control variables \cite{onorato2024bayesianoptimizationhyperparameterstuning}. The idea of BO is to construct a function surrogate of model's performance over hyperparameter space and use an acquisition strategy to sequentially select hyperparameters which lead to improved model performance. This intelligent selection strategy deliver strong sample efficiency relative to uninformed search, but often incurring high wall-clock cost for compute-intensive models \cite{shahriari2015taking}. Among BO variants, the Tree-structured Parzen Estimator replaces GP surrogates with density estimation over ``good" and "bad" observations and is widely adopted in practical frameworks for HPT \cite{watanabe2023treestructuredparzenestimatorunderstanding}. Further, BO-based HPT methods typically do not assume task-specific prior over the hyperparameter space, thus re-exploration of well-understood regions lead to redundant evaluations, specifically when historical configurations are available for common model–dataset pairs \cite{shahriari2015taking}.

\subsection{AutoML}
AutoML systems automate many parts of the machine learning pipeline. These include preprocessing, feature engineering, model selection, training, and HPT. They often rely on BO or grid search under standardized constraints. The goal is to produce deployable models with minimal manual intervention \cite{automl2019}. While accessible, this automation tend to behave as a black-box that limits interpretability and fine-grained control. Due to focus on general applicability, these may yield sub-optimal results compared to carefully engineered, domain-specific workflows under specialized resource or accuracy requirements \cite{hutter2019automated,automl2019}.

\begin{figure*}[t]
    \centering
    \includegraphics[width=0.8\textwidth]{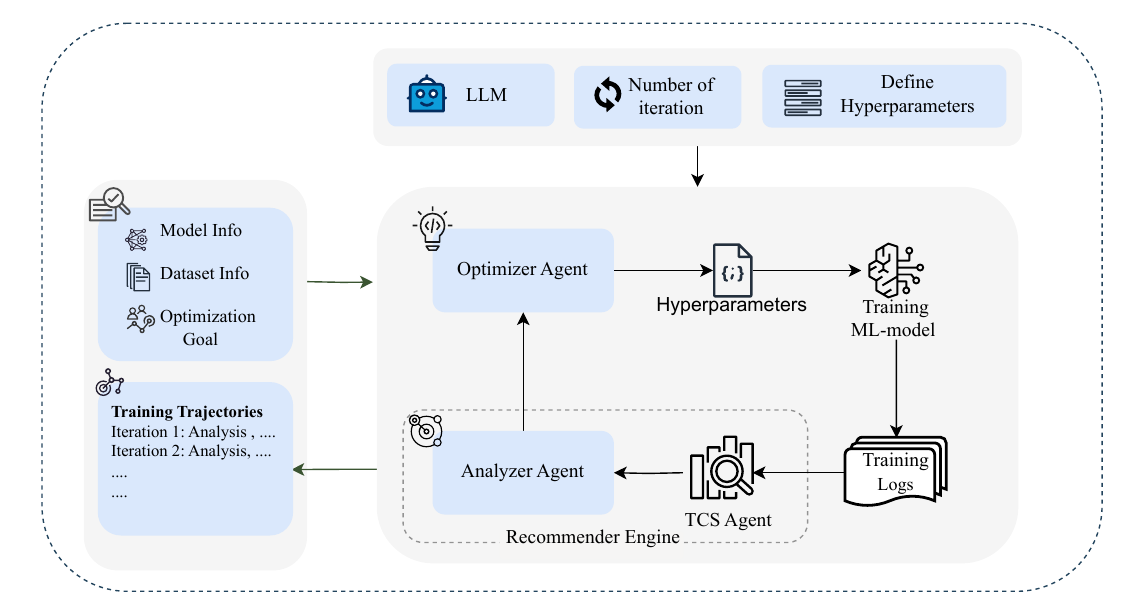}  
    \vspace{-1.5em}
    \caption{Flowchart of the proposed HPO pipeline with modular architecture.}
    \vspace{-1.5em}
    \label{fig:pipeline}
\end{figure*}
 
\subsection{Large Language Models for HPT}
 Recent reviews of LLM reasoning and agentic pipelines \cite{bubeck2023sparks,yao2022react}, highlight the potential of LLMs to act as iterative decision-makers. These LLM based agents can plan, observe feedback, and adapt. This has also motivated their use for HPT \cite{AgentHPO,zhang2023automl}. Recent AgentHPO work operationalizes this idea by having an LLM agent propose configurations, observe trial outcomes, and iterate, reporting improvements over baseline optimizers under comparable budgets across multiple tasks \cite{AgentHPO,zhang2023automl}. Complementarily, authors also show that iterative LLM-guided proposals can match or surpass traditional methods in accuracy per number of trials, strengthening evidence that LLMs can effectively steer HPT with limited evaluations \cite{zhang2024usinglargelanguagemodels,zhang2023mlcopilot}. A practical concern in LLM for HPT literature is reliance on \textit{very large proprietary models} (e.g., GPT-class systems), which raises questions about computational efficiency relative to traditional methods that explore broad parameter spaces like BO or random search \cite{AgentHPO,zhang2024usinglargelanguagemodels}.

Recently, authors in \cite{kochnev2025optunavscodellama} provides direct evidence that a fine-tuned, smaller open-source LLM can achieve—or surpass—state-of-the-art methods such as Tree-structured Parzen Estimator \cite{watanabe2023treestructuredparzenestimatorunderstanding}, while accelerating the tuning process, aligning the efficacy narrative with a resource-efficiency perspective. Also, studies in software engineering demonstrate that the performance of compact open-source language models (e.g., Llama-3.1-8B) on domain-specific tasks can be substantially enhanced without resorting to full fine-tuning. This improvement is achieved through the joint application of hyperparameter optimization and prompt engineering. To address the complexity of the search space, the approach employs the NSGA-II multi-objective optimization algorithm to efficiently prune candidate configurations, followed by a grid search procedure to identify Pareto-optimal solutions. The resulting configurations are evaluated with respect to both syntactic quality (e.g., fluency, grammaticality) and semantic quality (e.g., accuracy, contextual relevance) across multiple domains \cite{mahammadli2025sequentiallargelanguagemodelbased,madiraju2025optimindtunemultiagentframeworkintelligent}.

\paragraph{Positioning:}
In this work, our goal is to evaluate whether \textit{small LLMs can perform hyperparameter tuning and achieve performance comparable to state-of-the-art Bayesian Optimization, AutoML, or large LLM-based methods}. To answer this question, we design an agentic pipeline that includes a deterministic expert block acting as a context summarizer for the HPT task. Our approach builds on recent findings that (a) LLMs can drive effective HPT \cite{AgentHPO}, and (b) small open-source models can match larger proprietary ones when supported with structured optimization and prompting \cite{zhang2024usinglargelanguagemodels,mahammadli2025sequentiallargelanguagemodelbased}. The proposed expert-block framework equips small LLMs with deterministic, task-aware context from training trajectories, allowing them to analyze optimization progress with reliability under a fixed trial budget. Unlike studies focused mainly on very large models or domain-specific generation, our method targets general HPT with small LLMs by introducing a Trajectory Context Summarizer (TCS) that transforms raw training dynamics into structured inputs, improving proposal quality which reduces the analysis load on small LLMs, used in the proposed HPT pipeline.

\section{Proposed Methodology}
The proposed idea hinges on the fact that small LLMs struggle with reliable reasoning, especially when it requires multiple stages of analysis and a larger context window. In the case of the HPT problem, the LLM agent must dissect the epoch–loss curve from previous iterations to suggest the next hyperparameter setting, while also analyzing all possible HPT iteration trajectories. This is analogous to an optimization solver having access to gradients. To address this, we propose a deterministic, HPT-specific Trajectory Context Summarizer (TCS), designed to help small LLMs analyze the most recent trajectory and retain what occurred in prior HPT iterations.

Figure~\ref{fig:pipeline} illustrates our proposed HPT framework and its key components. At its core, the framework is designed to simplify and humanize the hyperparameter tuning process. Unlike conventional systems that rely solely on rigid mathematical modeling, our approach allows researchers to describe their model, dataset, and optimization goals directly in natural language. With this flexibility—combined with full control over the choice of LLM and the number of iterations—the framework becomes both powerful and easy to use.
The framework is organized into two modular, interacting components: the \textbf{Optimizer Agent} and the \textbf{Recommender Engine}. The Recommender Engine is central to the loop. It first restructures raw training logs into a reasoning-oriented format, enabling even smaller LLMs to operate with the depth and precision usually associated with larger models. This distilled information is then passed to an Analysis Agent, which guides the LLM to produce a structured, task-specific analysis by breaking the reasoning process into clear, manageable steps.

The refined insights are subsequently delivered to the Optimizer Agent, which generates improved hyperparameter candidates for the next iteration. This iterative cycle continues until the user-specified optimization rounds are completed.
By combining interpretability, modular reasoning, and problem-specific guidance, the framework gives users not only transparency but also control—whether it’s selecting the backbone LLM, setting iteration limits, or specifying detailed model and dataset information. The result is a human-centered, accessible, yet rigorous environment for hyperparameter optimization across diverse machine learning tasks. 

Following subsection presents the details of the optimizer agent, including our motivation and design of the prompt. Later, we present recommender engine, along with details of proposed TCS agent and Analyzer Agent.

\subsection{The Optimizer Agent}
\textbf{Optimizer Agent} is a small LLM tasked with generating and refining hyperparameters to improve performance of ML model. The core of our approach is a specific prompting strategy that frames hyperparameter tuning as an iterative, data-driven reasoning problem. The prompt is dynamically constructed at each optimization step to provide the agent with a comprehensive snapshot of the training progress, enabling it to make informed, directive-based decisions.

\begin{algorithm*}[b]
\caption{TCS-Enabled HPT}
\label{alg:ashpo}
\DontPrintSemicolon

\KwIn{
    $LLM_O$: Optimizer agent model; 
    $LLM_A$: Analysis agent model; 
    $M$: model architecture;
    $D$: dataset; \\
    $\mathcal{H}_{\text{space}}$: hyperparameter search space; 
    $G$: optimization goal; 
    $T$: maximum iterations.
}
\KwOut{
    $\Theta^*$: optimal hyperparameter configuration; 
    $\mathcal{L}$: complete experimental logs.
}

\BlankLine
\textbf{Initialization:} \\
$\text{Optimizer} \gets \textsc{InitializeOptimizer}(LLM_O, \mathcal{H}_{\text{space}}, G)$ \; \\
$\text{Recommender} \gets \textsc{InitializeRecommender}(LLM_A, G)$ \; \\
$\mathcal{L} \gets \emptyset$ \tcp*{Initialize experimental logs} \; 
$A_0 \gets \textsc{Recommender.BootstrapAnalysis}(\mathcal{H}_{\text{space}})$ \;

\BlankLine
\textbf{Main Loop:} \\
\For{$t \gets 1$ \KwTo $T$}{
    $(\Theta_t, J_t) \gets \textsc{Optimizer.Generate}(A_{t-1})$ \;\\
    $R_t \gets \textsc{ExecuteTraining}(M, D, \Theta_t)$ \;\\
    $\mathcal{L} \gets \mathcal{L} \cup \{(\Theta_t, J_t, R_t)\}$ \;\\
    $S_t \gets \textsc{Recommender.TCS.Summarize}(\mathcal{L})$ \;\\
    $A_t \gets \textsc{Recommender.AnalysisAgent.Analyze}(S_t)$ \;
}
\BlankLine
\textbf{Finalization:} \\
$\Theta^* \gets \textsc{FindBestConfiguration}(\mathcal{L}, G)$ \;\\
\Return $(\Theta^*, \mathcal{L})$ \;
\end{algorithm*}

\subsubsection{Prompt Architecture for Iterative Refinement:}
The agent's decision-making is guided by a multi-part prompt that acts as a form of \textbf{cognitive scaffolding}-- integrating real-time feedback, expert analysis, and a constrained action space. Each component is designed to focus the LLM on the singular goal of optimizing the target metric. The prompt structure is composed of the following critical elements:

\begin{itemize}
    \item \textbf{System-Level Persona and Goal:} We first establish the LLM's role as a machine learning expert. The system prompt explicitly defines the optimization objective, including the target metric (\texttt{metric\_name}), the desired direction of improvement (\texttt{target\_direction}), and a precise target value (\texttt{target\_value}). This instruction grounds the agent in the specific task of HPTx.

    \item \textbf{Dynamic Performance Context:} The user prompt provides immediate and relevant data from the latest training run, including a restated \textbf{Optimization Goal} and a quantitative summary of \textbf{Current Performance} (e.g., the current metric value, gap to target, and performance trend). This creates a direct \textbf{feedback loop} from the training environment to the agent.

    \item \textbf{Guided Analytical Input:} A crucial element of the prompt is the \textbf{Latest Analysis} section. This provides a concise, high-level recommendation derived from the training logs. The agent is explicitly instructed to base its decision \textit{strictly} on this analysis, which guides its reasoning and ensures its suggestions are hindged on analysis.

    \item \textbf{Constrained Search Space:} The prompt clearly delineates the \textbf{Hyperparameter Configuration}, separating parameters into fixed and optimizable sets. For each hyperparameter marked ``to optimize,'' we provide a valid range or a discrete set of options. This strictly defines the agent's action space, ensuring all generated configurations are valid.

    \item \textbf{Structured Task and Response Format:} Finally, the agent is given a precise set of instructions under \textbf{Your Task} and is mandated to follow a \textbf{structured output} format. This consists of two parts: a brief textual \textbf{reasoning} that justifies its decision based on the provided analysis, and a machine-readable \textbf{hyperparameters} string (e.g., \texttt{learning\_rate=0.005}).
\end{itemize}

Our main contribution is that this highly structured, feedback-driven prompt architecture can help the general-purpose LLM to focus on the  specialized task of optimization. It effectively leverages the model's reasoning capabilities while imposing the necessary constraints to ensure its outputs are reliable i.e. avoid unexplainable sudden changes, and programmatically parsable for our automated HPT loop.

\begin{figure*}[t]
    \centering
    \includegraphics[width=0.8\textwidth]{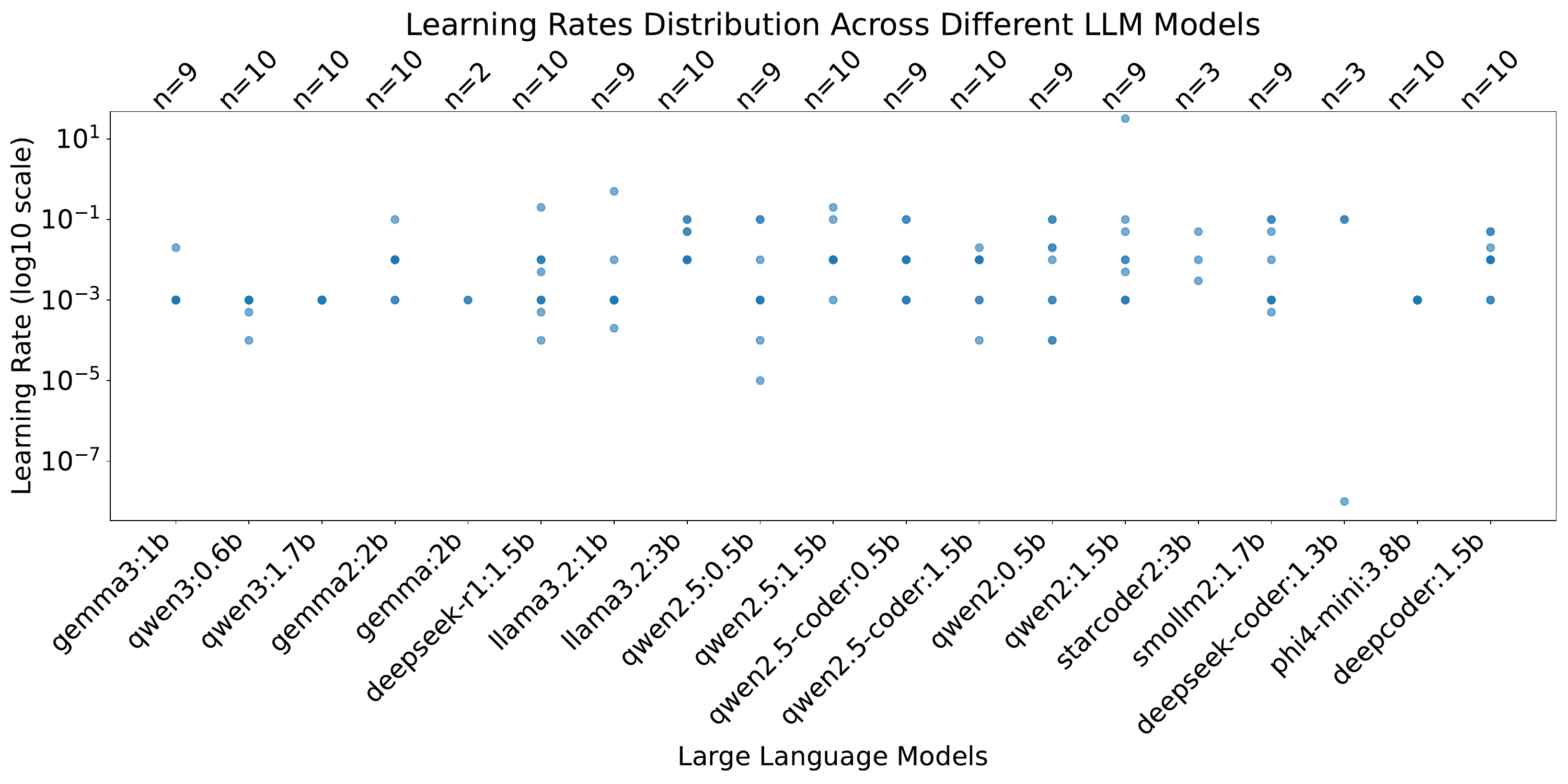}
    \vspace{-0.3em}
    \caption{Variation in learning rates proposed by small LLMs over repeated trials with the same input prompt. Each dot represents one valid response. The value of $n$ indicates the number of learning rates extracted after filtering invalid or irrelevant outputs. The wide spread illustrates the inconsistency of smaller models under identical conditions.}
    \label{fig:lr_fig}
    \vspace{-1.8em}
\end{figure*}

\subsection{The Recommender Engine}
The core of our HPT pipeline's intelligence resides in the Recommender Engine, a component responsible for analyzing the results of a training iteration and producing a reasoned, expert-level recommendation for the next set of hyperparameters. This engine is specifically designed to overcome a common limitation of smaller LLMs: a tendency to hallucinate or produce superficial analysis when presented with large, unstructured contexts like raw training logs. 

To mitigate this, the Recommender Engine employs a two-agent, modular architecture. The first agent, the \textbf{TCS (Trajectory Context Summarizer)}, distills the entire training history into a structured summary. The second, the \textbf{Analysis Agent}, then performs a guided analysis on this summary to generate a strategic recommendation. This output serves as the critical `analysis` input for the \textbf{Optimizer Agent} described in the previous section.

\subsection{Agent 1: Trajectory Context Summarizer (TCS)}
The motivation for TCS comes from the observation that small LLMs have limited reasoning capability, which leads to large variations in their predicted outputs, even when given the same input and prompt. We conducted a small experiment by asking for learning rates of a specific ML model, and as shown in Figure \ref{fig:lr_fig}, small LLMs fluctuated considerably.

\begin{figure*}[h]
\centering
    \vspace{1.2em}
    \includegraphics[width=0.9\linewidth]{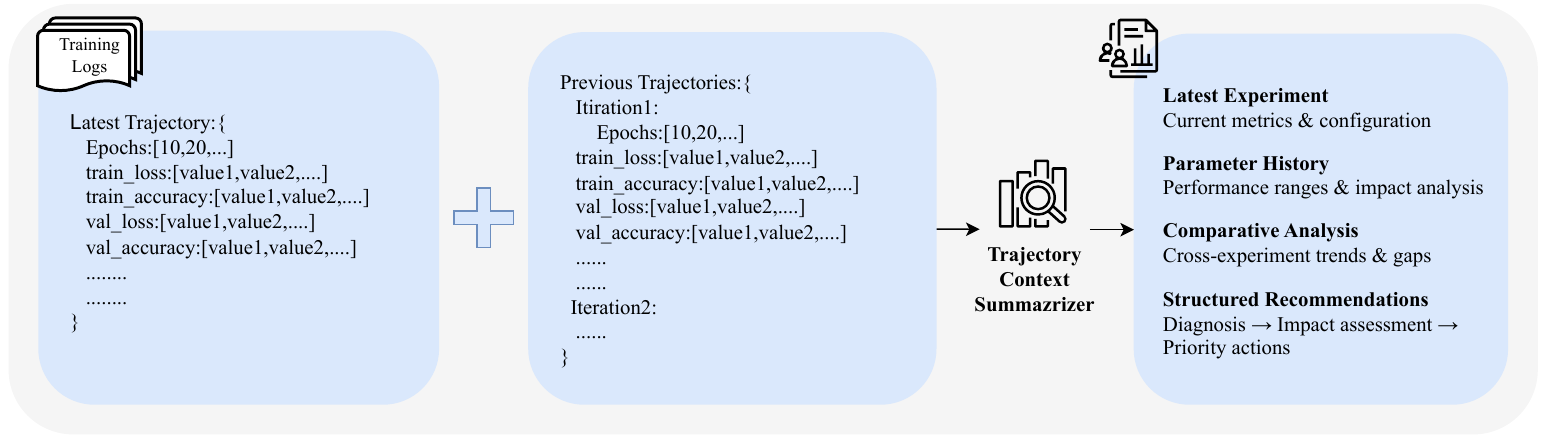}
    \caption{Conceptual depiction of input and output of the proposed Trajectory Context Summarizer.}
     \vspace{-1.2em}
    \label{fig:tcs}
\end{figure*}

The primary function of the Trajectory Context Summarizer (TCS) is to serve as a deterministic data preprocessor, analyzer  and context aggregator. After each training run, it collects and organizes all training related information into a comprehensive, structured report as shown in Figure \ref{fig:tcs}. This ensures that the subsequent analysis is grounded in factual data rather than abstract interpretation. The generated summary is a compact snapshot containing following key components:

\begin{itemize}
    \item \textbf{Current Situation:} A high-level overview of the optimization goal, the current metric 
    value, the gap to the user defined target, and the performance trend (e.g., \emph{IMPROVING}, \emph{STAGNATING}).

    \item \textbf{Latest Experiment:} The specific hyperparameters used in the most recent iteration 
    and the resulting performance metrics.

    \item \textbf{Detailed Hyperparameter Analysis:} A parameter-by-parameter breakdown of the entire experiment history. For each hyperparameter, it lists its valid range, its current value, and its historical performance, while flagging unexplored regions.

    \item \textbf{Previous Experiment Comparison:} A differential analysis between the latest and 
    previous runs, highlighting the hyperparameter changes and their impact on the target metric.
\end{itemize}

\subsection{Agent 2: Analysis Agent}

The Analysis Agent is a small LLM tasked with performing a root-cause analysis of HPT task based on completed iterations and formulating the next step. Its performance comes not from a single complex prompt, but from a step-by-step support method, often called cognitive scaffolding. In our case, this means using two instructional templates that guide the LLM’s reasoning in a structured way.

First, a system-level directive establishes the agent's persona as an expert optimizer and provides it with a mandatory reasoning approach. This framework compels the agent to follow a specific diagnostic sequence:

\begin{enumerate}
    \item Identify the primary problem like overfitting or underfitting etc.
    \item Assess which hyperparameter categories (e.g., regularization, model capacity) which are most likely to address the problem identified in step 1.
    \item Select a \textit{single most impactful} hyperparameter to adjust.
    \item Provide clear cause-and-effect reasoning for the choice of hyperparameter in step 3.
\end{enumerate}

Second, the user-facing directive takes the structured summary from system level directive and embeds it within a rigorous analytical framework. This framework requires the agent to populate a six-part response:

\begin{enumerate}
    \item \textbf{Problem Diagnosis:} A concise statement of the core issue.
    \item \textbf{Hyperparameter Impact Assessment:} A theoretical evaluation of how each available parameter could influence the metric.
    \item \textbf{Primary Action:} The selection of the single most critical hyperparameter to modify.
    \item \textbf{Specific Recommendation:} The exact new value or direction of change for the chosen parameter.
    \item \textbf{Reasoning:} A detailed justification for the recommendation over other alternatives.
    \item \textbf{Expected Outcome:} A prediction of the anticipated performance improvement.
\end{enumerate}

This two-part approach—first giving a factual summary and then guiding a step-by-step reasoning process—helps the small LLM break down the complex task of hyperparameter tuning into smaller, manageable steps. Our idea is that asking small LLMs to do this analysis in two different ways can reduce the randomness in their responses.

\subsection{Complete TCS-Enabled HPT Loop}


The proposed HPT cycle has following flow. Each iteration $t$ begins with the \textbf{Recommender Engine} providing a strategic report, $A_{t-1}$, based on the historical log. This analysis directs the \textbf{Optimizer Agent} to generate a new set of hyperparameters, $\Theta_t$, and a corresponding justification, $J_t$. The system then executes a training run with $\Theta_t$ to yield performance results, $R_t$. The pipeline for the iteration is completed as the \textbf{Recommender Engine} takes over:  \textbf{TCS} first processes the results ($R_t$) to create a structured summary, $S_t$. This summary is then passed to the \textbf{Analysis Agent}, which produces the final strategic report, $A_t$. This report, $A_t$, then serves as the primary directive to kickstart the next optimization cycle. This continuous loop of analysis, generation, and execution creates a powerful feedback mechanism that systematically steers the search toward $\Theta^*$. Details of the same are also given in Algorithm \ref{alg:ashpo}.

\begin{figure*}[t]
    \centering
    \begin{subfigure}[b]{0.24\textwidth}
        \centering
        \includegraphics[width=\textwidth]{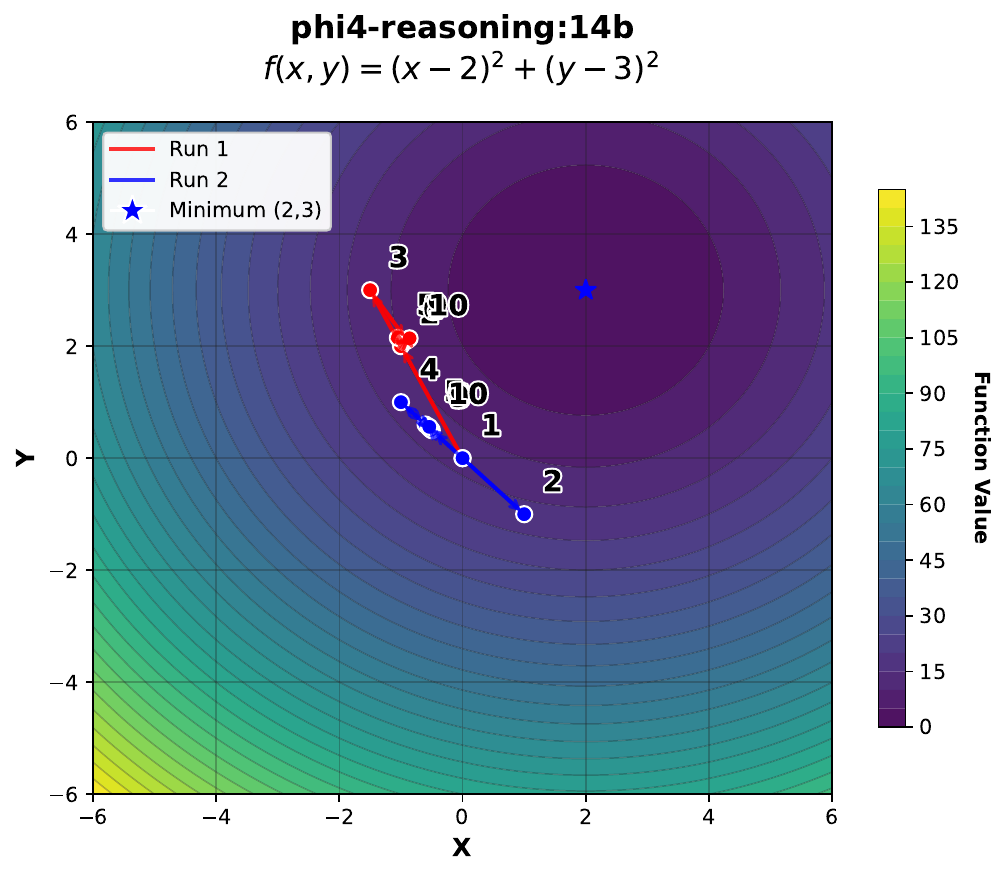}
        \caption{Phi-3.5 Reasoning (14B) \\ without TCS}
        \label{fig:phi_gpt_normal_pdf}
    \end{subfigure}
    \hfill
    \begin{subfigure}[b]{0.24\textwidth}
        \centering
        \includegraphics[width=\textwidth]{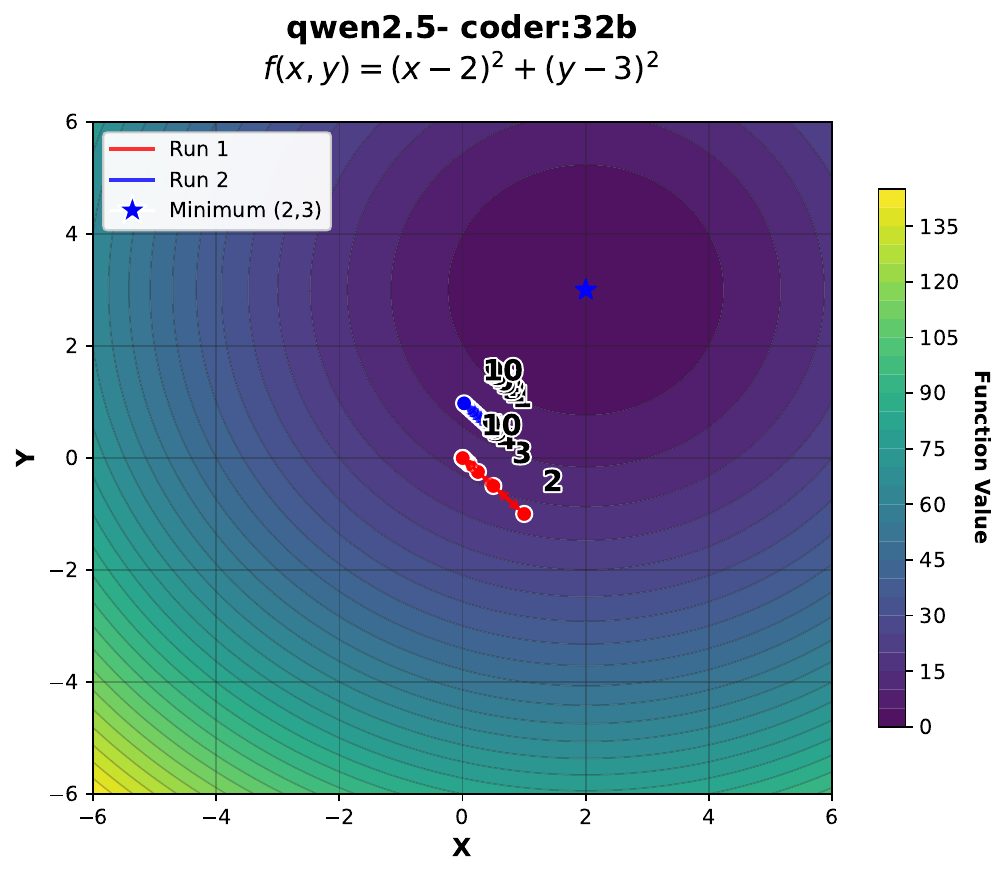}
        \caption{Qwen2.5-Coder (32B) \\ without TCS}
        \label{fig:coder_gpt_normal_pdf}
    \end{subfigure}
    \hfill
    \begin{subfigure}[b]{0.24\textwidth}
        \centering
\includegraphics[width=\textwidth]{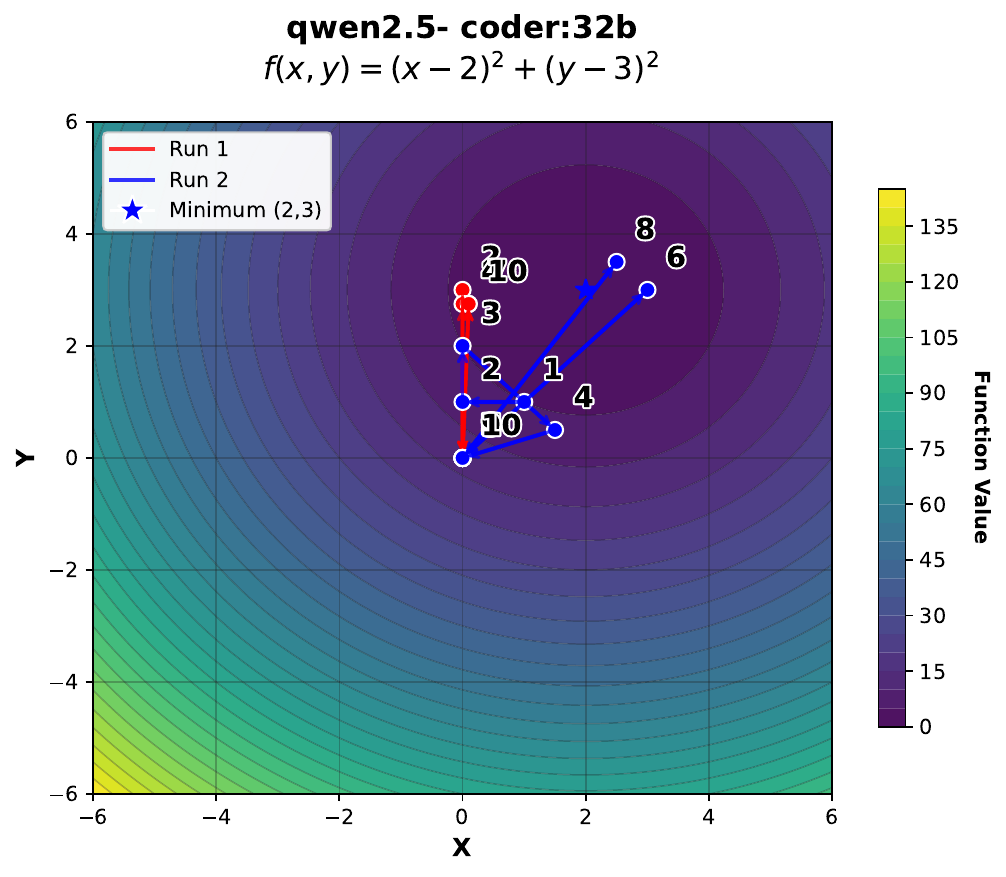}
        \caption{Qwen2.5-Coder (32B) \\ 
        with TCS}
        \label{fig:coder_gpt_advanced_pdf}
    \end{subfigure}
    \hfill
    \begin{subfigure}[b]{0.24\textwidth}
        \centering
        \includegraphics[width=\textwidth]{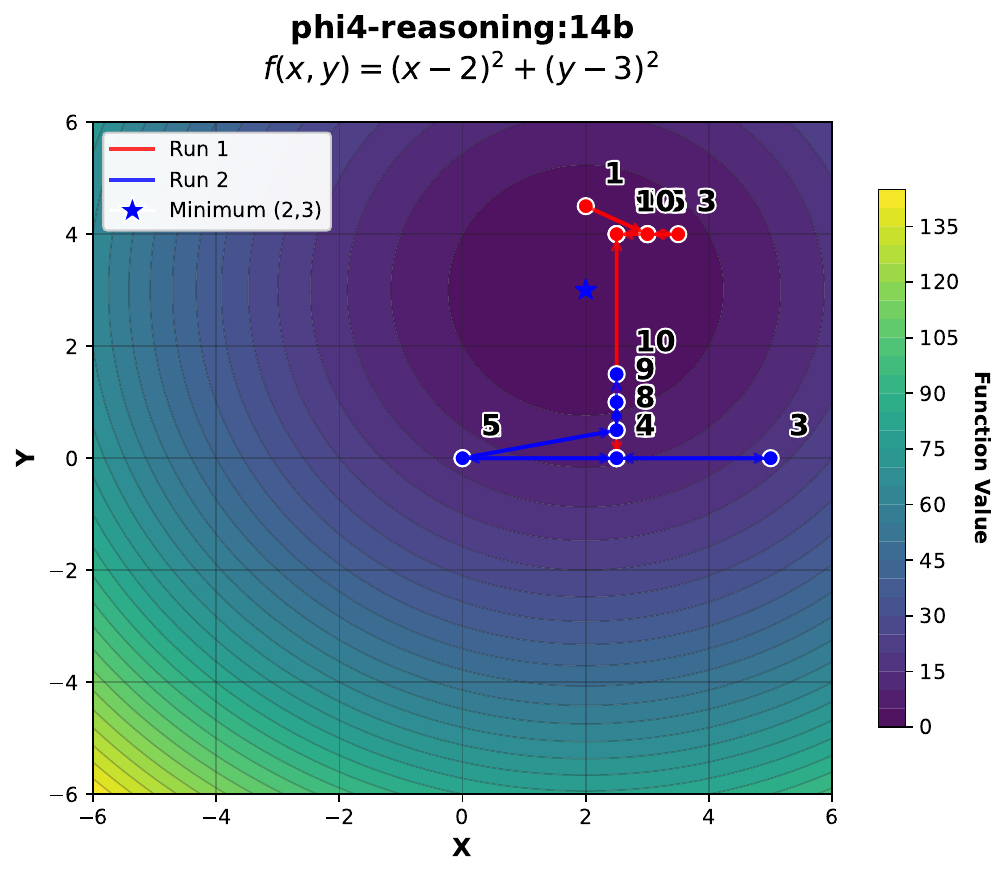}
        \caption{Phi-4 Reasoning (14B) \\ with TCS}
        \label{fig:phi_gpt_advanced_pdf}
    \end{subfigure}
    \vspace{-0.5em}
    \caption{Optimization trajectories for different LLMs with and without the Trajectory Context Summarizer (TCS). 
    The left two plots show runs without TCS, where both models fail to converge efficiently. 
    The right two plots show runs with TCS, where structured summaries enable both models to converge reliably to the global minimum.}
    \label{fig:combined_tcs_comparison}
\end{figure*}

\section{Results and Discussions}
The proposed methodology was applied to six distinct tasks spanning five key domains: Computer Vision (CV), Natural Language Processing (NLP), Recommender Systems (RecSys), Tabular Learning, and Graph Neural Networks (GNN). The task selection was deliberately curated to include both foundational benchmarks (e.g., CIFAR-10, Cora) and contemporary, real-world problems to ensure a robust assessment of our framework's capabilities. A detailed summary of the domains, applications, datasets, target models, and primary evaluation metrics is presented in Table~\ref{tab:domain_performance} in Appendix. To demonstrate the effectiveness of the proposed TCS-enabled HPT tool, we present a comparative analysis with the study in \cite{AgentHPO}. This choice aligns with our goal of evaluating whether the proposed expert system block can help reduce the performance gap between small LLM-based HPT pipelines and proprietary large LLM-based tools. Some important implementation details and language models used are discussed below.

The proposed framework leverages the reasoning capabilities of small, locally-run language models deployed via the \textbf{Ollama} framework \cite{ollama2025}. The two primary models used for the Optimizer and Recommender agents were \texttt{phi4:reasoning14b} and \texttt{qwen2.5-coder:32b}. This approach ensures reproducibility and control over the experimental environment. To ensure the reliability and stability of our results, we conducted \textbf{5 independent HPT runs} for each task. Each run consisted of \textbf{10 iterations (trials)}. The final reported performance is the mean of the best performance achieved across these 5 independent runs. All experiments are done on a system with Intel Xeon W7-2495X processor having 256GB RAM and Nvidia RTX 4500Ada graphics card. Implementation details, code, prompts and scripts can be found at \url{https://github.com/PSquare-Lab/LLM_TCS_HPT}


Below we first discuss inconsistency in small LLM responses, then analyses the effectiveness of TCS using a function optimization problem and then discuss HPT results as a comparative analysis with results from \cite{AgentHPO}.

\subsection{Inconsistency in Responses of Small LLMs}
We tested the reasoning ability of small LLMs by repeatedly giving them the same prompt. Each model was asked ten times to propose the best learning rate for improving accuracy given a set of training trajectories. As shown in Figure~\ref{fig:lr_fig}, the responses of small LLMs were highly inconsistent. Models with fewer parameters produced scattered answers and suggested very different learning rates across trials. In some cases, the models ignored the instructions and generated irrelevant outputs that could not be plotted. When the model size increased, the variability in responses decreased. Larger models converged to a smaller set of learning rate values and showed more reproducibility across repeated prompts. This result highlights a key limitation of small LLMs: their lack of consistency makes them unreliable for tasks like hyperparameter selection. Increasing model size reduces this issue and leads to more stable reasoning.

\begin{table*}[h]
\vspace{-1em}
\centering
\small
\caption{Performance comparison across different tasks. All values are mean $\pm$ std\%.}
\vspace{-1em}
\label{tab:performance_comparison_tiny_pm}
\resizebox{\textwidth}{!}{%
\begin{tabular}{l|ccccc|cc}
\toprule
\textbf{Task} & \textbf{Random} & \textbf{Bayesian} & \textbf{GPT-3.5} & \textbf{GPT-4} & \textbf{Human} & \textbf{Proposed Method} & \textbf{Without TCS} \\
\midrule
Tabular Regression & 56.49 {\scriptsize $\pm$0.40\%} & 56.85 {\scriptsize $\pm$0.05\%} & 56.78 {\scriptsize $\pm$0.36\%} & 58.01 {\scriptsize $\pm$0.11\%} & 56.9 & 56.82 {\scriptsize $\pm$0.52\%} & 56.02 {\scriptsize $\pm$0.52\%} \\
Node Classification & 78.80 {\scriptsize $\pm$0.81\%} & 74.96 {\scriptsize $\pm$2.70\%} & 81.13 {\scriptsize $\pm$0.22\%} & 81.38 {\scriptsize $\pm$0.22\%} & 81.5 & 80.87 {\scriptsize $\pm$0.19\%} & 75.48 {\scriptsize $\pm$1.52\%} \\
Image Classification (CIFAR-10) & 81.63 {\scriptsize $\pm$3.88\%} & 79.87 {\scriptsize $\pm$4.19\%} & 83.87 {\scriptsize $\pm$1.18\%} & 85.18 {\scriptsize $\pm$0.52\%} & 85.05 & 84.75 {\scriptsize $\pm$0.99\%} & 81.17 {\scriptsize $\pm$3.91\%} \\
Image Classification (Butterfly) & 78.99 {\scriptsize $\pm$2.32\%} & 63.57 {\scriptsize $\pm$8.67\%} & 84.79 {\scriptsize $\pm$1.01\%} & 85.92 {\scriptsize $\pm$0.57\%} & 82.74 & 87.00 {\scriptsize $\pm$2.15\%} & 69.70 {\scriptsize $\pm$3.55\%} \\
Text Classification (SST-2) & 90.28 {\scriptsize $\pm$0.76\%} & 90.28 {\scriptsize $\pm$0.45\%} & 90.34 {\scriptsize $\pm$0.79\%} & 91.32 {\scriptsize $\pm$0.11\%} & 90.71 & 89.58 {\scriptsize $\pm$0.43\%} & 87.48 {\scriptsize $\pm$18.31\%} \\
Click Through Rate (CTR) & 81.94 {\scriptsize $\pm$0.24\%} & 81.67 {\scriptsize $\pm$0.92\%} & 82.14 {\scriptsize $\pm$0.06\%} & 82.09 {\scriptsize $\pm$0.05\%} & 82.19 & 82.03 {\scriptsize $\pm$0.02\%} & 80.66 {\scriptsize $\pm$0.01\%} \\
\bottomrule
\end{tabular}%
}
\vspace{-1.5em}
\end{table*}

\subsection{Effect of TCS: Study using Function Optimization}
Before presenting large-scale benchmarks, we first study a simple minimization problem $(x-2)^2 + (y-3)^2$ to test whether the Trajectory Context Summarizer (TCS) improves optimization. This task serves as a controlled setting where the impact of structured versus unstructured optimization signals can be clearly observed.

In the baseline setup, raw training logs are passed directly to the analysis agent without TCS. Figure~\ref{fig:combined_tcs_comparison} shows that this leads to unstable optimization across different models. The Phi-3.5 model oscillates around suboptimal points and does not show consistent progress. The Qwen2.5-Coder model, despite having more parameters, also fails to converge and follows a random-walk pattern. Even GPT-3.5, used as a commercial reference, does not reach the optimum under this setup. These outcomes suggest that without structured preprocessing, LLMs—small or large—struggle to interpret training trajectories, retain memory across trials, or systematically guide the search.

When TCS is introduced, the optimization behavior changes significantly. As shown in Figure~\ref{fig:combined_tcs_comparison}, both models now display clear and directed trajectories toward the global minimum. The Qwen2.5-Coder model systematically explores the space, avoiding previously tested regions and moving closer to the optimum in each step. The Phi-3.5 reasoning model, though smaller, achieves faster convergence with fewer iterations. This demonstrates that reasoning-focused LLMs, when supplied with structured trajectory context, can outperform larger models in efficiency.  

The improvement can be explained by the role of TCS in restructuring the optimization history. Instead of presenting the LLMs with long, noisy logs, TCS condenses the information into a state-like summary. This allows the analysis agent to:  maintain optimization memory across iterations, compare new results against past trials, Identify unexplored hyperparameter ranges and Propose next steps based on structured evidence rather than raw signals.  This experiment shows that the presence of TCS is the deciding factor in whether small LLMs can perform reliable hyperparameter tuning. Without TCS, even large or commercial models fail to converge. With TCS, smaller open-source models produce stable and efficient optimization trajectories. These findings validate our design choice and motivate the use of TCS in the full benchmarking experiments that follow.

\begin{figure*}[t]
    \centering
    \includegraphics[width=0.9\textwidth]{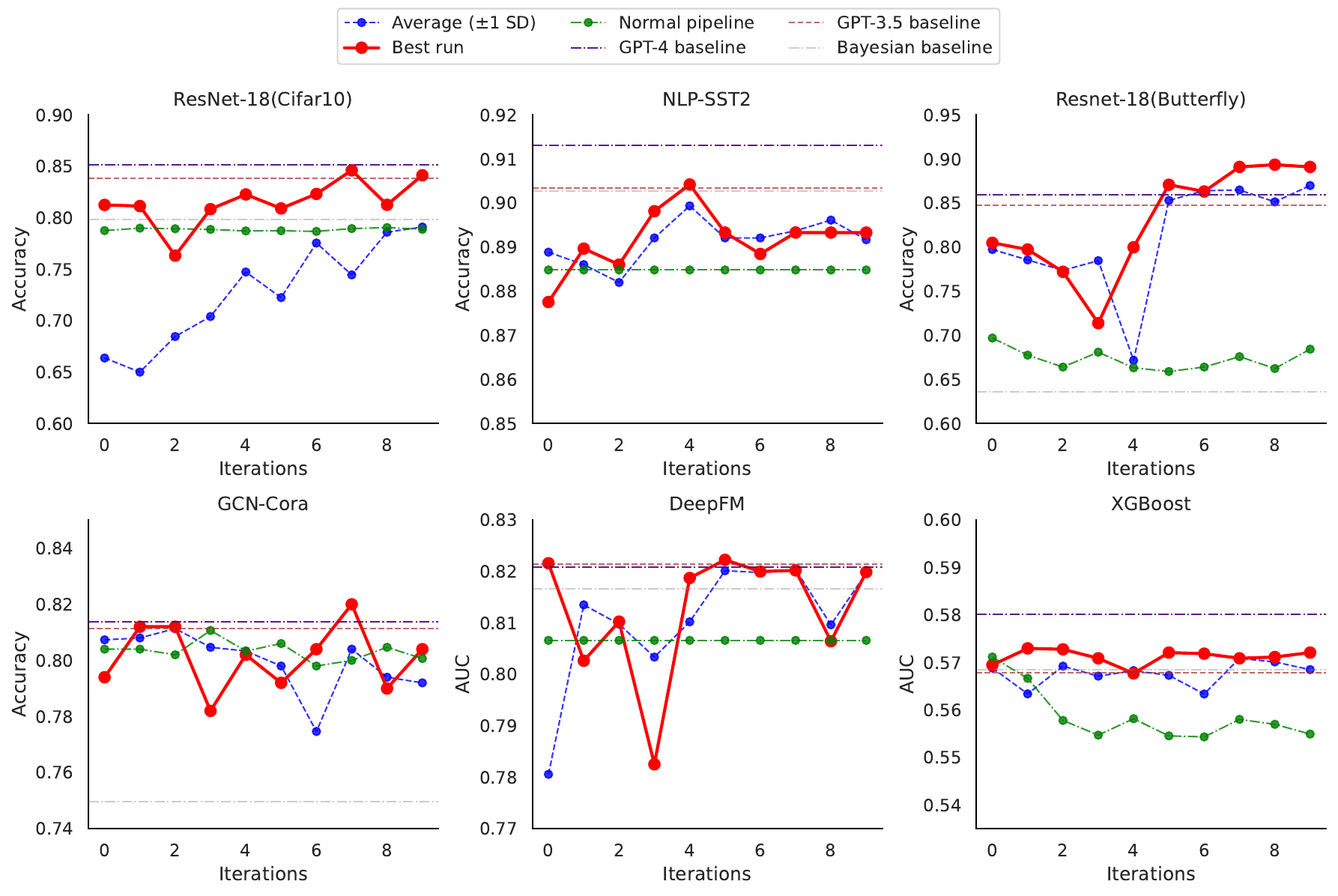}  
    \caption{\textbf{Phi4:reasoning14b-Instruct Performance.} Results are averaged over 5 independent runs. Our AS-HPO pipeline (blue line with shaded SD) significantly outperforms our Normal Pipeline (green line, an ablation without the Summarizer) and the GPT-3.5 baseline, while performing competitively with the GPT-4 baseline.}
    \label{fig:phi4_reasoning}
     \centering
    \includegraphics[width=0.9\textwidth]{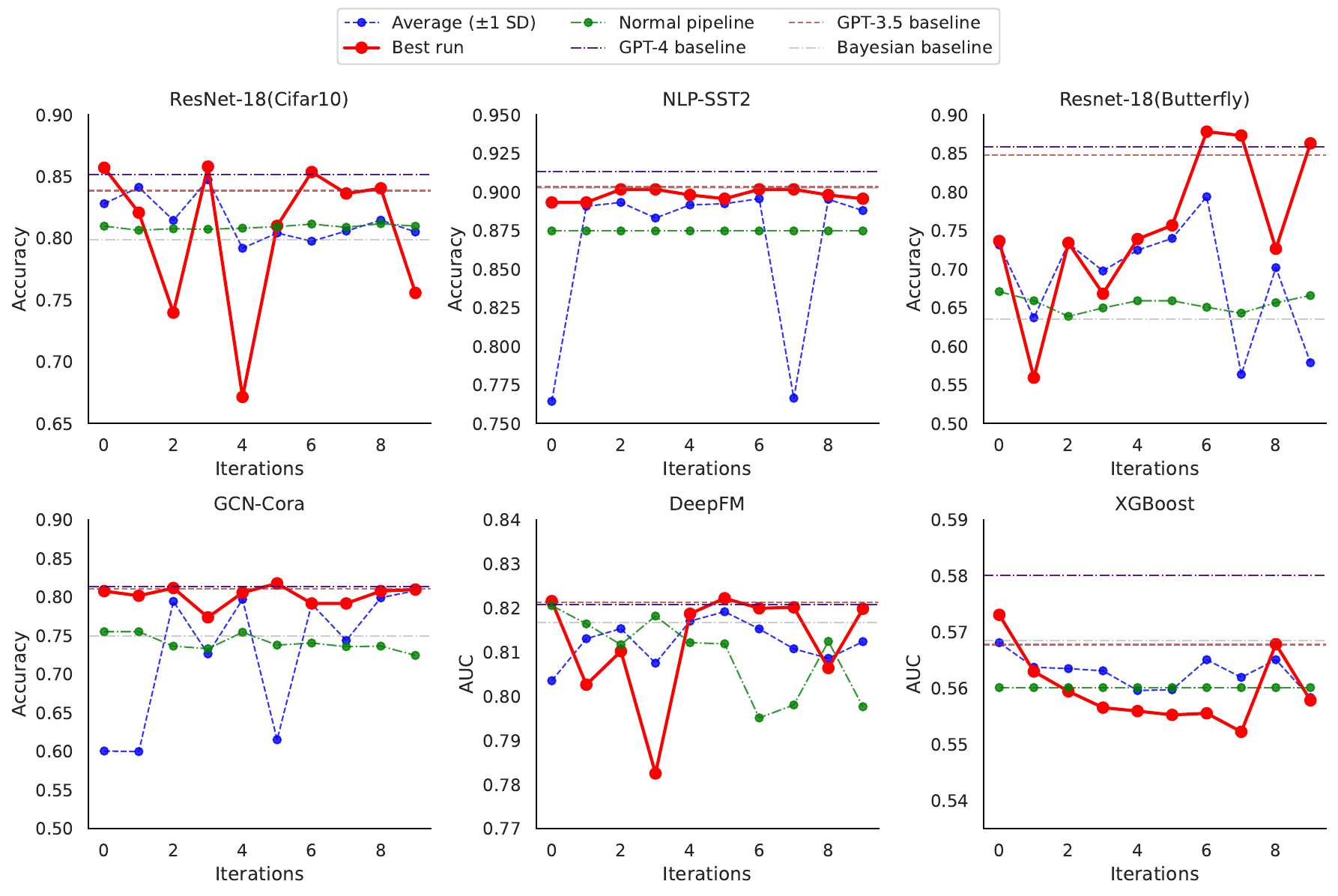}  
    \caption{\textbf{Qwen2.5 Coder-32B Performance.} Results are averaged over 5 independent runs. The AS-HPO pipeline (blue line) demonstrates strong performance, surpassing the GPT-3.5 and Bayesian baselines and showing competitive results against GPT-4, validating our approach with a different model architecture.}
    \label{fig:qwen_coder}
\end{figure*}

\subsection{Results for HPT Task}
As explained before, our target is to see if TCS can elevate small LLM up to large LLM level performance for HPT task. So, for a fair and consistent comparison, all benchmark values for baselines  results are taken directly from the results reported in the original \textbf{AgentHPO} paper \cite{AgentHPO}. Our evaluation compares proposed TCS-Enabled HPT against two categories of methods:

\begin{enumerate}
    \item \textbf{Traditional HPT Methods:} We measure our framework's efficiency and performance against two widely-recognized traditional methods: \textbf{Random Search}, which remains a formidable baseline \cite{bergstra2012random, HPObench}, and \textbf{Bayesian Optimization}, a prominent model-based approach.
    
    \item \textbf{State-of-the-Art LLM Optimizers:} To situate our work at the forefront of this field, we directly compare our small-model approach against the performance of leading large, proprietary models: \textbf{GPT-3.5} and \textbf{GPT-4}.
\end{enumerate}

For the task of HPT, we first conducted experiments using small LLMs (up to 3.8B parameters). These models exhibited a lack of consistency in their responses, making it difficult to extract hyperparameters reliably. A major limitation was hallucination, where the models often generated incomplete or misleading instructions. From these observations, it became evident that increasing model size improved response stability, with larger models producing more consistent outputs for identical inputs.

Based on these findings, we subsequently tested moderately larger models and observed that certain ones provided stable and reproducible results. In particular, \texttt{Qwen2.5-32B} \cite{qwencoder} and \texttt{Phi-4-14B} \cite{phireson} were evaluated in depth. To assess their effectiveness, we developed a two-agent pipeline consisting of an  Optimizer Agent and a Recommender Engine, which iteratively refined hyperparameters using training trajectories. Two pipeline variants were tested: one incorporating a TCS to preprocess trajectories, and one without it (raw training trajectories passed directly to the analyzer). 


\subsection{Detailed Performance Analysis}

We now analyze performance across six representative tasks, comparing our \textbf{Proposed Method (TCS-Enabled HPT)} against the baseline pipelines and existing methods. Graphical representation of these results are given in Figure \ref{fig:phi4_reasoning} for \texttt{Phi-4-14B} model and Figure \ref{fig:qwen_coder} for \texttt{Qwen2.5-32B} model. Also, results of multiple methods with standard deviation across trials are given in Table \ref{tab:performance_comparison_tiny_pm}

\textit{\textbf{Task 1: Image Classification (ResNet-18 on CIFAR-10)}}  
On CIFAR-10, the Proposed Method improves accuracy over the GPT-3.5 baseline and performs closely to GPT-4. The baseline pipeline without TCS shows stagnation and limited exploration. This highlights that structured trajectory summaries allow small models to converge more reliably on vision tasks.  

\textit{\textbf{Task 2: Text Classification (NLP-SST2)}}  
For SST-2, the Proposed Method surpasses GPT-3.5 and achieves results comparable to GPT-4 across runs. The pipeline without TCS again fails to adapt, producing flat trajectories. This suggests that TCS is important for handling the longer training histories typical in NLP tasks.  

\textit{\textbf{Task 3: Image Classification (ResNet-18 on Butterfly Dataset)}}  
On the Butterfly dataset, the Proposed Method reaches accuracy levels close to GPT-4, while the pipeline without TCS trends downward and fails to find stable improvements. The contrast indicates that structured context helps avoid poor updates that can drive the model away from optimal regions.  

\textit{\textbf{Task 4: Node Classification (GCN on Cora)}}  
In the GCN task, the Proposed Method exceeds GPT-3.5 and remains close to GPT-4, while the pipeline without TCS shows little improvement. This demonstrates that the approach generalizes beyond vision and text, supporting reasoning in graph-based learning tasks.  

\textit{\textbf{Task 5: Click-Through Rate Prediction (DeepFM)}}  
For the recommender system task, the Proposed Method improves AUC steadily and reaches levels on par with GPT-4. Without TCS, the baseline pipeline struggles to progress, showing that structured summaries are necessary for complex interaction features in tabular models.  

\textit{\textbf{Task 6: Tabular Classification (XGBoost)}}  
On the XGBoost benchmark, the Proposed Method again performs above GPT-3.5 and remains competitive with Bayesian Optimization and GPT-4. The pipeline without TCS remains flat, confirming that TCS enables the agent to make use of trial history when optimizing simple but sensitive tabular models.

Across all six tasks, the proposed method consistently improves over the baseline without TCS and produces results competitive with large LLM baselines. The improvements are most pronounced where training logs are long or noisy, supporting the role of TCS in compressing trajectories into usable state information.

\section{Conclusion and Future Work}

This work introduced a trajectory–summarization-based framework for hyperparameter tuning, designed to enable small, open-source LLMs to operate effectively under constrained trial budgets. By integrating the Trajectory Context Summarizer (TCS), we showed that small LLMs can deliver results competitive with Bayesian Optimization, AutoML approaches, and even larger commercial LLMs, while avoiding the instability observed in unstructured pipelines. Our analysis across diverse tasks—including vision, text, tabular, graph, and recommendation domains—demonstrates that structured summarization and guided reasoning are essential for reliable performance in hyperparameter optimization. In future work, we aim to extend the framework in three directions. First, by incorporating richer forms of summarization, such as uncertainty-aware or probabilistic representations of trajectories. Second, by exploring the integration of reinforcement learning signals into the agent loop, enabling adaptive exploration–exploitation balancing. Finally, we plan to evaluate the method on large-scale, industrial datasets and under distributed multi-agent settings, where communication efficiency and robustness to partial information will become critical. Together, these directions will further clarify the potential of small LLMs as practical and resource-efficient optimizers.

\begin{table*}[p]
\centering
\caption{Domain-Specific Model Summary}
\label{tab:domain_performance}
\renewcommand{\arraystretch}{1.2}
\setlength{\tabcolsep}{3pt}
\footnotesize
\begin{tabularx}{\textwidth}{c c c c c c}
\toprule
\textbf{Field} & \textbf{Application} & \textbf{Data Source} & \textbf{Best Model} & \textbf{Hyperparameters [Range]} & \textbf{Measure} \\
\midrule

\multirow{1}{*}{\textit{Computer Vision}} 
& \begin{tabular}{@{}c@{}}Image \\[-2.5pt] Classification \end{tabular} & \begin{tabular}{@{}c@{}}CIFAR-10 \\ Butterfly Image \end{tabular} & ResNet-18 &
\begin{tabular}{@{}c@{}}
lr: [10$^{-5}$, 10$^{-1}$]; optimizer: [adam, sgd]; \\
epochs: [25, 200]; weight\_decay: [10$^{-6}$, 10$^{-1}$]; \\
dropout: [0, 0.5]; momentum: [0.5, 1]; \\
batch\_size: [32, 64, 128, 256]
\end{tabular} & Accuracy \\
\midrule

\multirow{1}{*}{\begin{tabular}{@{}c@{}}\textit{Natural Language}  \\[-2.5pt] \textit{Processing} \end{tabular}} 
& \begin{tabular}{@{}c@{}}
Text \\[-2.5pt] Classification \end{tabular}
& SST2 & DistilBERT &
\begin{tabular}{@{}c@{}}
lr: [10$^{-6}$, 10$^{-2}$]; epochs: [1, 4]; \\
dropout/attn/seq\_dropout: [0, 0.5]; \\
batch\_size: [8, 16, 32, 64, 128]; \\
activation: [gelu, relu, silu]; weight\_decay: [10$^{-6}$, 0.1]
\end{tabular} & Accuracy \\
\midrule

\multirow{1}{*}{\begin{tabular}{@{}c@{}}\textit{Recommender}  \\[-2.5pt] \textit{Systems} \end{tabular}} 
& \begin{tabular}{@{}c@{}}
Click-Through \\[-2.5pt] Rate \end{tabular} & Movie Lens & DeepFM &
\begin{tabular}{@{}c@{}}
lr: [10$^{-5}$, 10$^{-1}$]; embed\_size: [8, 16, 32, 64]; \\
optimizer: [adam, sgd]; reg\_weight: [10$^{-6}$, 10$^{-1}$]; \\
dropout: [0, 0.5]; batch\_size: [256, 512, 1024, 2048]; \\
mlp\_hidden: [32, 64, 128, 256, 512]; mlp\_layers: [1, 4]
\end{tabular} & AUC \\
\midrule

\multirow{1}{*}{\textit{Tabular Learning}} 
& Regression & House Price & XGBoost &
\begin{tabular}{@{}c@{}}
lr: [10$^{-3}$, 1]; max\_depth: [3, 11]; \\
min\_child\_weight: [1, 10]; subsample: [0.5, 1]; \\
colsample\_bytree: [0.5, 1]; n\_estimators: [100, 500]; \\
gamma: [0, 0.5]; reg\_alpha/reg\_lambda: [0, 1]; \\
scale\_pos\_weight: [1, 10]
\end{tabular} & R² Score \\
\midrule

\multirow{1}{*}{\begin{tabular}{@{}c@{}}\textit{Graph Neural}  \\[-2.5pt] \textit{Networks} \end{tabular}} 
& \begin{tabular}{@{}c@{}}Node \\[-2.5pt] Classification \end{tabular} & \begin{tabular}{@{}c@{}}Cora Citation  \\[-2.5pt] Network \end{tabular} & GCN &
\begin{tabular}{@{}c@{}}
layers: [1, 5]; lr: [10$^{-6}$, 10$^{-1}$]; optimizer: [adam, sgd]; \\
epochs: [1, 200]; hidden\_size: [8, 16, 32, 64]; \\
activation: [relu, elu, silu]; weight\_decay: [10$^{-6}$, 10$^{-1}$]; \\
dropout: [0, 0.5]
\end{tabular} & Accuracy \\
\bottomrule
\end{tabularx}
\end{table*}

\section{Acknowledgments}
The authors would like to thank Dr. Abhijith Jayakumar (LANL) for his valuable initial discussions on the use of LLMs for HPT.

\bibliography{main}


\end{document}